\documentclass[letterpaper, 10 pt, conference]{ieeeconf}  

\IEEEoverridecommandlockouts                              

\usepackage{graphics} 
\usepackage{epsfig} 
\usepackage{mathptmx} 
\usepackage{times} 
\usepackage{amsmath} 
\usepackage{amssymb}  
\usepackage{amsfonts} 
\usepackage{lipsum}
\usepackage{cite}
\usepackage[T1]{fontenc}
\usepackage{float}
\usepackage{algpseudocode}
\usepackage{algorithm}
\usepackage[hidelinks]{hyperref}
\usepackage[usenames, dvipsnames]{color}

\title{\LARGE \bf
Wallbounce : Push wall to navigate with Contact-Implicit MPC 
}

\author{Xiaohan Liu$^{1*}$, Cunxi Dai$^{1*}$, John Z. Zhang$^{1}$, Arun Bishop$^{1}$, Zachary Manchester$^{1}$ and Ralph Hollis$^{1}$  
\thanks{*Equal contribution.}
\thanks{$^{1}$Robotics Institute, Carnegie Mellon University, Pittsburgh, PA 15213, USA, {\tt\small\{xiaohan5, cunxid, ziyangz3, arunbish, zmanches, rhollis\}@cs.cmu.edu}}%
}

\begin{document}

\maketitle
\thispagestyle{empty}
\pagestyle{empty}

\begin{abstract}


In this work, we introduce a framework that enables highly maneuverable locomotion using non-periodic contacts. This task is challenging for traditional optimization and planning methods to handle due to difficulties in specifying contact mode sequences in real-time. To address this, we use a bi-level contact-implicit planner and hybrid model predictive controller to draft and execute a motion plan. We investigate how this method allows us to plan arm contact events on the shmoobot, a smaller ballbot, which uses an inverse mouse-ball drive to achieve dynamic balancing with a low number of actuators. Through multiple experiments we show how the arms allow for acceleration, deceleration and dynamic obstacle avoidance that are not achievable with the mouse-ball drive alone. This demonstrates how a holistic approach to locomotion can increase the control authority of unique robot morpohologies without additional hardware by leveraging robot arms that are typically used only for manipulation. Project website: \href{https://cmushmoobot.github.io/Wallbounce/}{\textcolor{blue}{https://cmushmoobot.github.io/Wallbounce}} 



\end{abstract}

\section{INTRODUCTION}
Humans and animals possess the incredible natural ability to leverage interactions between all parts of their bodies and the environment to achieve highly agile and dynamic locomotion behaviors. These interactions enable them to increase their control authority and locomotion capabilities beyond what can be achieved with legs alone. For example, parkour athletes use their hands to push off against walls and navigate around obstacles. Inspired by these capabilities, solving complicated locomotion tasks by leveraging diverse contact sources has been a long-standing challenge in robotics research.


Existing literature on multi-contact motion planning and control largely considers locomotion \cite{perspectivelocomotion}\cite{qpmpc} and manipulation \cite{osti_5761101} as separate research problems. In recent years, the rising interest in generalist robot agents has accelerated the design of robot hardware platforms equipped with both wheeled bases or legs for locomotion and arms for manipulation \cite{ballbot7dof}, \cite{fu2024mobilealoha},\cite{anymal}. This new trend in robot morphology also introduces interesting research questions on how one can take advantage of the addition of robot arms during locomotion to augment the capabilities and robustness of the robot. Despite these increasingly mature human-like robot form factors, systems capable of solving challenging locomotion problems while leveraging upper-body capabilities remain understudied.

Integrating upper limb contacts into a Model Predictive Control (MPC) framework, however, is challenging. MPC often requires a predefined contact schedule for each contact point. For legged locomotion, where contacts are typically periodic, a contact schedule can be generated with heuristics~\cite{qpmpc,anymal}. For upper limb contacts, determining the timing and duration of contact for other parts of the body is difficult. Most of the existing work rely on a hand-crafted contact schedule for upper limbs~\cite{CentroidalPlanning,ETHMPCWBC}. Other works use search-based methods to find possible contact strategies \cite{ETHLSTP,DarpaAcyclic}. However, these methods are based on kinematics and quasi-static analysis and cannot capture the dynamic effect of the robot. 

\begin{figure}[t!]
    \centering    \includegraphics[width=\columnwidth]{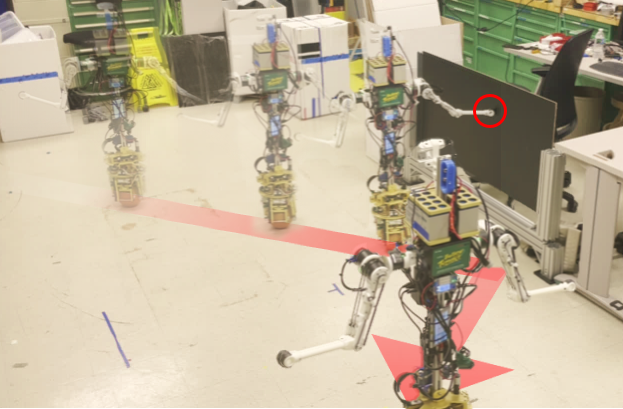}
    \caption{Time-lapse picture of CMU shmoobot making a sharp turn by pushing wall.}
    \label{fig:cover}
    \centering
    \vspace*{-3mm}
\end{figure}

In this paper, we investigate using end effector contact on the CMU shmoobot (a smaller CMU ballbot \cite{nagarajan2014ballbot}) to enhance its balance, locomotion, and navigation capabilities. Shmoobot uses a single ball drive to achieve dynamic balancing with a low number of actuators while remaining maneuverable in tight human spaces. However, this unique morphology limits how quickly Shmoobot can change its momentum, making it less robust when encountering unexpected obstacles while moving or experiencing large disturbances. We explore how we can use the arms the robot would typically use for manipulation to address these limitations during locomotion, increasing shmoobot's robustness and reactivity with no additional hardware.

We propose a bi-level Model-Predictive Control (MPC) framework that enables our robot to discover and utilize upper limb contacts during locomotion. At the higher level, we use contact-implicit optimization to identify potential contact schedules. Then, at a lower level, we deploy a hybrid trajectory optimization with this fixed contact schedule to generate smooth, feasible motion plans. Finally, we implement this framework on the CMU shmoobot platform and demonstrate its capabilities through several hardware experiments.

Our specific contributions are:

\begin{enumerate}
    \item A bi-level MPC framework that can reason about acyclic contacts and leverage upper limb contacts in locomotion.
    \item Deployment and evaluation of the proposed framework on a novel bi-manual service robot that balances on a ball.
    \item Experiments demonstrating how upper limb contacts can effectively assist in robot locomotion.
\end{enumerate}





\section{Related Works}\label{sec: relatedworks}


\subsection{Locomotion With Upper Limb Contacts}
Like humans, human-like robots can use their upper limbs to assist with locomotion. Research~\cite{wang_realization_2018} explored how a humanoid robot can make contact with a wall to prevent falling, while~\cite{DarpaAcyclic, CentroidalPlanning} demonstrated how arm contact can help robots traverse challenging terrains. More related research studied how robots can gain acceleration by making contact with the environment. Reference~\cite{HumanPushing} focuses on transferring human wall-pushing skills directly to robots. In \cite{PushingNavigation}, the researcher developed a reflex-based controller that can control the moving direction of a robot after contact with a wall. Our work focuses on an optimization-based framework that enables the CMU shmoobot to autonomously leverage upper limb contact without pre-specification during locomotion and navigation.

\subsection{Contact-Implicit  Optimization}
Optimization-based algorithms can be a powerful tool when planning over contact mode schedules and timings. In particular, contact-implicit optimization (CIO)\cite{manchester2019contact, posa_direct_2014} does not require predefined contact timings or locations, allowing the algorithm to explore different contact patterns. One common method for solving CIO is to formulate contact dynamics as complementarity problems\cite{LCP}. Research~\cite{Hopper2016} treats the contact dynamics as constraints and solves them with the direct collocation method. A higher-order collocation method~\cite{StagedContactOptimiztion, manchester2019contact}, solves the problem using HTO to refine the solution for faster convergence. However, due to the non-smooth nature of the contact dynamics, these methods are numerically unstable and take a long time (minutes to hours) to converge~\cite{zhang2023slomo}, making them unpractical during deployment in online MPC settings without specialized solvers~\cite{cleach2024fast}. Another approach of solving CIO is by using compliant contact models. Works~\cite{CIOManipulation, DiscoverBehaviorCI02019}, use a continuous contact flag to control the allowed contact force. Other works~\cite{RALNonlinearSpring,RALNonlinearSpring2}, use a nonlinear spring-damper system to model contact and solve the CIO problem with DDP-based methods. Compliant contact models make fast, real time CIO possible. However, soft contact models often introduce physical artifacts like force at a distance, which makes the planned trajectory hard to track for the hardware.









\section{Background}\label{sec: background}

\subsection{Platform Description}

The CMU shmoobot (Fig.~\ref{fig:PlatformDescription}(a)) is a 1.2 m tall robot that balances on a ball wheel. The robot has a pair of {3-DOF} torque-controllable arms mounted onto its body. The ball is actuated by a four-motor Inverse Mouse-Ball Drive mechanism (IMBD)\cite{nagarajan2014ballbot}. A pair of actuated opposing rollers drive the ball in each of the two orthogonal motion directions, which allows omnidirectional motion on the floor. A slip ring assembly and 5th actuator allows unlimited yaw rotation of the body. The model makes the following assumptions: (i) there is no slip between the ball and the floor; and (ii) the ball is always in contact with the floor.  

\subsection{Symbol and Notations}

The body and world coordinate systems are defined in Fig.~\ref{fig:PlatformDescription}. Quantities in the body frame have a left subscript $B$. Other quantities are in the world coordinate system. Vectors are bold and lowercase ($\mathbf{a}$, $\omega$), matrices are uppercase ($A$, $\Omega$), scalars are lowercase and italicized ($\mathit{a}$, $\mathit{\omega}$). 

The operator $[{\mathbf{v}}]_{\times}$ converts a vector $\mathbf{v}=\left[v_1 ; v_2 ; v_3\right] \in \mathbb{R}^3$ into a skew-symmetric 'cross product matrix':
\begin{equation}
[{\mathbf{v}}]_{\times}=\left[\begin{array}{ccc}0 & -v_3 & v_2 \\ v_3 & 0 & -v_1 \\ -v_2 & v_1 & 0\end{array}\right],
\end{equation}
and we have $\mathbf{v} \times \mathbf{x} = [{\mathbf{v}}]_{\times} \mathbf{x}$.

\subsection{MPC overview}

Consider that we have an optimal control problem with $I$ modes:

\begin{equation}
\label{eq:OCProblem}
    \begin{cases}
        \underset{\mathbf u(.) \ x(.)}{\min} \ \ \sum_i \Phi_i(\mathbf x(t_{i+1})) + \displaystyle \int_{t_i}^{t_{i+1}} l_i(\mathbf x(t), \mathbf u(t), t) \, dt \\
        \text{s.t.} \ \mathbf x(t_0) = \mathbf x_0 & \hspace{-7em} \text{(\ref{eq:OCProblem} - 1)} \\
        \dot{\mathbf x}(t) = \mathbf f(\mathbf x(t), \mathbf u(t), t) &\hspace{-7em}\text{(\ref{eq:OCProblem} - 2)} \\
        {\mathbf g}_i(\mathbf x(t), \mathbf u(t), t) = \mathbf{0} &\hspace{-7em} \text{(\ref{eq:OCProblem} - 3)} \\
        \mathbf h_i(\mathbf x(t), \mathbf u(t), t) \geq \mathbf{0} &\hspace{-7em} \text{(\ref{eq:OCProblem} - 4)} \\
        \text{for  } t_i < t < t_{i+1} \text{  and  } i \in \{0, 1, \cdots, I-1 \}
    \end{cases}.
\end{equation}

Here, (\ref{eq:OCProblem}-1) is the initial state constraint; (\ref{eq:OCProblem}-2) is the system dynamics constraint; (\ref{eq:OCProblem}-3) and (\ref{eq:OCProblem}-4) are equality and inequality constraints respectively. In our formulation, the active constraints vary as the mode of the system changes. An MPC controller recurrently solves this optimal control problem and searches for an optimal state input trajectory that minimizes the overall stage cost.

\section{System Modeling}\label{sec: system_model}

\begin{figure}[t]
    \centering
    \includegraphics[width=\columnwidth]{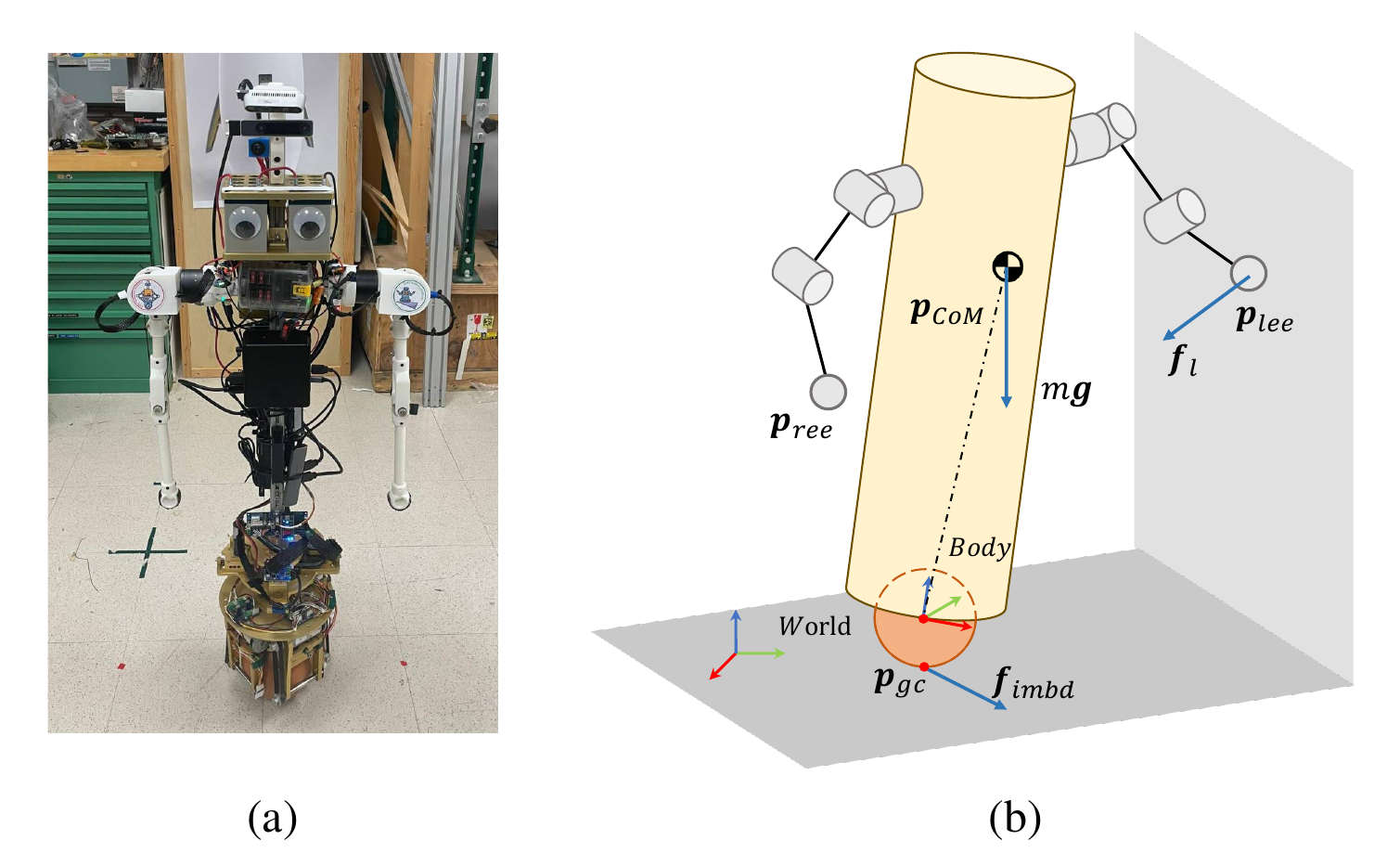}
    \vspace{-6mm}
    \caption{(a) CMU shmoobot. 
    (b) Coordinate systems of Shmoobot.}
    \label{fig:PlatformDescription}
    \centering
    \vspace*{-4mm}
\end{figure}

\subsection{ Coordinate Definition}
Since the ball is always in contact with the ground, we have a set of position constraints $ p_z^{ball} = r_{ball}, v_z^{ball} = 0$ and $ a_z^{ball} = 0$. The system dynamics is modeled with a set of minimal coordinates with implicit dynamics constraints.

As shown in Fig.~\ref{fig:PlatformDescription}(b), the origin of the robot base frame is defined at the center of the ball wheel. The general coordinate of the base is then defined as $\mathbf q_b = [p_x^{ball}, p_y^{ball}, \psi, \theta, \phi]^T$, where $p_x^{ball}, p_y^{ball}$ are position, $\psi, \theta, \phi$ are ZYX Euler angles of body orientation.

\subsection{Simplified Robot Dynamics}

We use single rigid body dynamics to model the shmoobot system. We neglect the mass of the arms and only consider the dynamic effect of the robot body.

Since the base frame is not defined at the center of mass, we need to consider the dynamics coupling between angular and linear terms. From Newton-Euler equations we have:

\begin{equation}
\begin{aligned}
\sum \mathbf f^i = m\mathbf{\dot v} + m [{\mathbf r}_{com}]_{\times} \boldsymbol \alpha -m[{\boldsymbol \omega}]_{\times}[{\boldsymbol \omega}]_{\times} {\mathbf r}_{CoM},  \\
{\sum \boldsymbol \tau^i} + \sum [{\mathbf r_{f}^i}]_\times \mathbf f^i = \mathbf{\dot{l}}_{r}-  [{\mathbf r}_{com}]_{\times} m \mathbf a + [\mathbf{\omega}]_\times I_{inertia} \mathbf{\omega},
\label{eqn:NewtonEuler}
\end{aligned}
\end{equation}
where $m$ is the body mass, $I_{inertia}$ is the inertia matrix, $\mathbf{v}$ is the linear velocity; and $\mathbf{l_{r}}$ is the angular momentum. $\mathbf{f^i}$ and $\mathbf{\tau^i}$ are applied external force and torque respectively. ${\mathbf r}_{CoM}$ is the position vector from the origin of the base frame to the center of mass (CoM). Similarly, $\mathbf r_{f}^{i}$ is the position vector from the origin to the point where external force $\mathbf{f^i}$ is applied.

In our case, we assume that the system has small angular velocity and angular acceleration. Also, since the ball always remains on the ground, we have $\mathbf r_z = 0, \mathbf v_z = 0$ and $\mathbf a_z = 0$. Then the linear acceleration can be expressed as 
$\mathbf a = \frac{[\mathbf{\sum f}^i_x, \mathbf{\sum f}^i_y, 0]^T} {m}$.

Then we have:

\begin{gather} 
m\mathbf{\dot{v}} = \sum \mathbf [ \mathbf{f}^i_x, \mathbf{f}^i_y , 0]^T ,\\
\mathbf{\dot{l}_{r}} = {\sum \boldsymbol \tau^i} +   \sum [{\mathbf r_{f}^i}]_\times \mathbf f^i +  [{\mathbf r}_{}]_{\times} [\mathbf{\sum f^{i}_x}, \mathbf{\sum f^{i}_y}, 0]^T.
\label{eqn:SimplifiedDynamics}
\end{gather}

In our case, we only consider the contact force on the end effectors and the ball. Then, the base dynamics of the system can be written as:

\begin{gather}
m\mathbf{\dot{v}} =[\mathbf{f}_{IMBD, x},\mathbf{f}_{IMBD, y}, 0]^T + \sum_{i=1}^2\left[\mathbf{f}_{c,x}^i, \mathbf{f}_{c,y}^i, 0\right]^T, \\
\mathbf{\dot{l}_{r}} =\sum_{i=1}^2 \mathbf{r}_{b, c}^i \times \mathbf{f}_{c}^i + \mathbf{r}_{b,CoM} \times m \mathbf{g} +  \mathbf{\tau}_{yaw} + \mathbf{r}_{b,CoM} \times m\mathbf{\dot{v}}.
\label{eqn:ShmooRigidBodyDynamics}
\end{gather}

Here, $\mathbf{f}_{IMBD}$ is the contact force on the ball. $\mathbf{f}_{c}^i$ is the contact force on the i-th end effector. $\mathbf{r}_{b, c_i}$ is the position of the i-th end effector w.r.t. the center of the ball; $\mathbf{r}_{b, CoM}$ is the position vector from ball center to the center of mass.

\section{Contact Modeling}
\label{sec:ContactModeling}
In this section, we present the contact models used by the controllers. The Contact-Implicit MPC uses a contact-invariant soft-contact model, while the Hybrid MPC uses a linear constraints-based set of contact modes.

\subsection{Contact Frame Definition}
\begin{figure}[t!h]
    \centering
    \includegraphics[width=0.95\columnwidth]{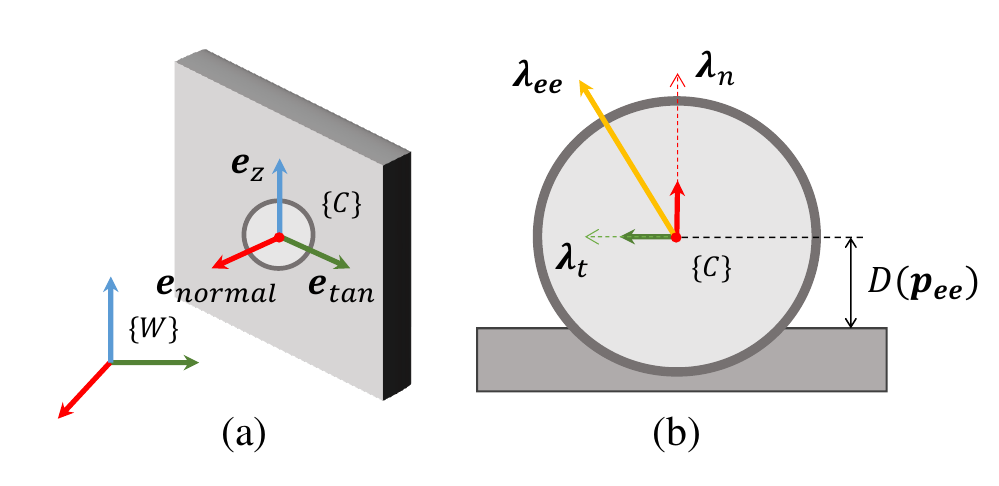}
    \vspace{-3mm}
    \caption{(a) Definition of contact frame. (b) Schematic of contact.}
    \label{fig:ContactDefinition}
    \centering
\end{figure}

As shown in Fig.~\ref{fig:ContactDefinition}(a), 
the orientation of the contact frame is determined by the contact surface. The surface normal $\mathbf{e_{\text{normal}}}$ at a given point $\mathbf{p}$ can be acquired by computing the gradient of the Signed Distance Function (SDF) of the surface:
\begin{equation}
\mathbf{e_{\text{normal}}} = \nabla D(\mathbf{p}).
\end{equation}

In this paper, we only consider vertical surfaces, so vector $\mathbf{e_z} = [0, 0, 1]^T$ is always tangent to the surface. Then, we can obtain the other tangent vector by taking the cross product:
\begin{equation}
    \label{eqn:normalForce}
    \mathbf{e}_{tangent} = \mathbf{e}_{normal} \times \mathbf{e}_{z} .
\end{equation}

\begin{figure*}[t]
    \centering
    \includegraphics[width=2\columnwidth]{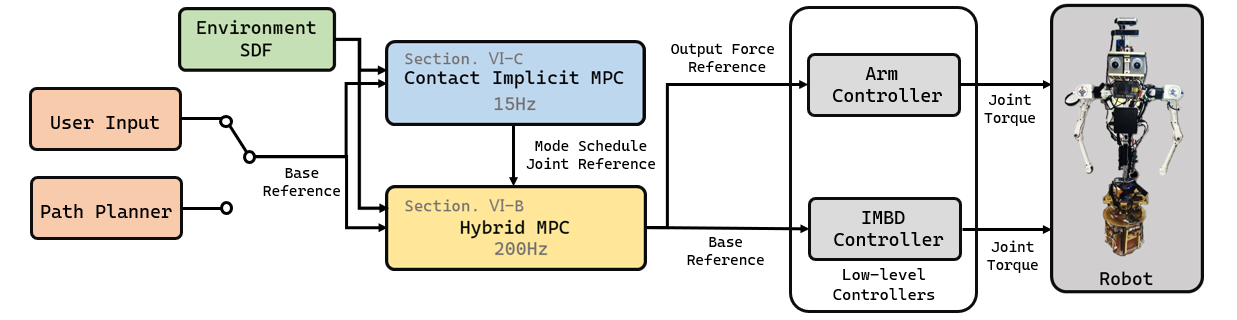}
    \caption{Control framework diagram. A reference trajectory is first generated by a path planner or sent by the user. A bi-level MPC will calculate an optimal trajectory that tracks the reference. The environment SDF used by the controller needs to be recomputed. The upper level of the controller (the blue block) is a contact-implicit MPC which generates a draft of the motion plan with soft contact models. The lower level of the controller (the yellow block) is a hybrid MPC. It will extract a contact schedule from the motion plan, and refine the trajectory with hard contact models. The low level balancing controller and arm controller will then track the motion plan provided by the hybrid MPC.}
    \label{fig:ControllerStrucutre}
    \centering
    \vspace*{-3mm}
\end{figure*}

\vspace{-3mm}

\subsection{Soft Contact Model}
\label{subsec:SoftContact}
In contact-implicit MPC, we used a contact-invariant soft contact model. The normal part of the contact force is expressed as a nonlinear function of end-effector position: 
\begin{equation}
\lambda_{normal}=f(D(\mathbf{p}_{ee})).
\label{eqn:SoftContactEqn}
\end{equation}
Here, $\mathbf{p}_{ee}$ is the position of the end effector, $D(\mathbf{p})$ is the signed distance function of the surface. $f(d)$  is a nonlinear scalar-valued function that increases rapidly when $d$ is smaller than 0, and sticks to 0 when $d$ is larger than 0. There are many choices of the activation function $f(d)$, here we pick 
\begin{equation}
f(d) = 0.5 f_{max} \cdot tanh(- \alpha \cdot (d+\beta)) + 0.5 f_{max},
\end{equation}
where $\alpha$ and $\beta$ controls the stiffness of the contact and $f_{max}$ is the maximum allowed normal force. This model implicitly contains the information of contact force limit and is twice differentiable.

The end effector should always be outside the surface, so we have:

\begin{equation}
    \label{eqn:posconstraint}
    D(\mathbf{p}_{ee}) \geq 0.
\end{equation}

Finally, the end effector shouldn't slip on the wall when the contact force is nonzero. This gives a linear complementary constraint:

\begin{equation}
    \label{eqn:velconstraint}
    \lambda_{\text{ee}} \dot{\mathbf{p}}_{\text{ee}} = 0.
\end{equation}

\begin{figure*}[htp]
    \centering
    \includegraphics[width=1.8\columnwidth]{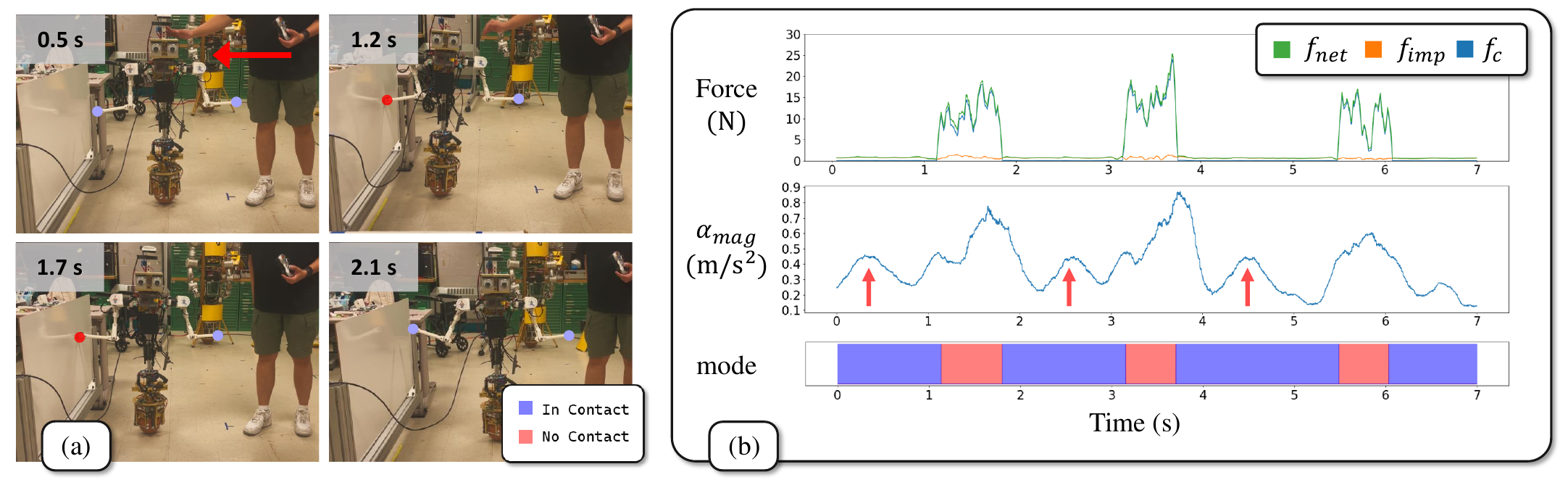}
    \caption{(a) Key frames for the disturbance rejection experiment. The end effectors are highlighted with blue and red circles. A red circle indicates that the end effector is in contact. At T = 0.5 s, a human operator pushed the robot to the wall. The robot actively used its arm to recover from the disturbance. (b) Force output trajectory, acceleration trajectory and mode sequence of the robot.}
\label{fig:DisturbanceRejectionTimeElapse}
    \centering
    \vspace*{-3mm}
\end{figure*}

\subsection{Hybrid Contact Model}
\label{subsec:HardContact}
For hybrid MPC, we use a hybrid contact model. We denote the set of closed contacts by $\mathrm{C}$. Then, if an end effector is in contact, we have the following constraints:

\begin{equation}
\label{eqn:ContactConstraint}
\begin{cases}
\dot{\mathbf{p}}^{\text{i}}_{ee} =\mathbf{0} \\
D(\mathbf{p}^{\text{i}}_{ee}) =\mathbf{0} \\
\mathbf{\lambda}^{\text{i}}_{normal} \cdot \mathbf{e}^{i}_{normal} > \mathbf{0}  \\
-\mu \mathbf{\lambda}_{normal}^i \leq \mathbf{\lambda}_{tan}^i \leq \mu \mathbf{\lambda}_{normal}^i \\
-\mu \mathbf{\lambda}_{normal}^i \leq \mathbf{\lambda}_{z}^i\leq \mu \mathbf{\lambda}_{normal}^i
\end{cases}
\text { if } c_i \in {\mathrm{C}} .
\end{equation}
If the end effector is not in contact, we have the following constraints:
\begin{equation}
\label{eqn:NoContactConstraints}
\begin{cases}
\lambda^{\text{i}}_{ee} =\mathbf{0} \\
D(\mathbf{p}^{\text{i}}_{ee}) \geq \mathbf{0} 
\end{cases}
\text { if } c_i \in \overline{\mathrm{C}}.
\end{equation}

The set of contacts, $\mathrm{C}$, is be obtained from the optimized trajectory of the contact-implicit controller. This will be further discussed in Section.~\ref{section:controller}.

\section{System Design}\label{sec: system_design}
\label{section:controller}

In this section, we present our locomotion controller framework. The overall system structure is presented in Fig.~\ref{fig:ControllerStrucutre}.

\subsection{Hybrid MPC}
\label{sec:hybridmpc}
\subsubsection{System Dynamics}
The system states $\mathbf{x} \in \mathbb{R}^{16}$ and inputs $\mathbf{u} \in \mathbb{R}^{15}$ are defined as:

\begin{equation}
\mathbf{x} = \left[\mathbf{h}_{b}^T, \mathbf{q}_{b}^{T},\mathbf{q}_{j}^{T}\right]^{T}, \mathbf{u} = \left[\mathbf{f}_{IMBD}^T,\mathbf{f}_{c}^T,\mathbf{v}_j^T\right]^T.
\end{equation}

Here, $\mathbf{q}_b$ is the generalized coordinate of the base. $\mathbf{q}_j$ are the joint positions. $\mathbf{h}_{b}=\left[m\mathbf{v}^T, \mathbf{l}_{r}^T\right]^T \in \mathbb{R}^5$ is the collection of linear and angular momentum. 

For input $\mathbf{u}$, $\mathbf{f}_{IMBD} = [\mathbf{f}_x, \mathbf{f}_y, \mathbf{\tau}_z]$ is the contact force on the ball. Due to the unique property of the IMBD mechanism, there is almost no relative spinning along the z axis of the ball, and we can approximately have $\mathbf{\tau}_z = \mathbf{\tau}_{yaw}$. $\mathbf{f}_{c} = [\mathbf{f}_{c}^1, \mathbf{f}_{c}^2] \in \mathbb{R}^{6}$ is the contact force on the two end effectors. $ \mathbf{v}_j^T$ are the joint velocities.

Then, from equations (\ref{eqn:ShmooRigidBodyDynamics}) we have:

\begin{equation}
\frac{\mathrm{d}}{\mathrm{d} t}\left[\begin{array}{c}
m\mathbf{v} \\
\mathbf{l_{r}} \\
\mathbf{q}_b \\
\mathbf{q}_j
\end{array}\right] = \\
\left[ \begin{array}{c}
\sum_{i=1}^2\left[\mathbf{f}_{c,x}^i, \mathbf{f}_{c,y}^i\right]^T + [\mathbf{f}_{IMBD, x},\mathbf{f}_{IMBD, y}]^T\\
\sum_{i=1}^2 [\mathbf{r}_{b, c}^i] _\times \mathbf{f}_{c}^i + [\mathbf{r}_{b, CoM}]_\times m (\mathbf{g}+\dot{\mathbf{v}}) + \mathbf{\tau}_{yaw} \\
A_b^{-1}(\mathbf q_b)\mathbf{h}_{b}\ \\
\mathbf{v}_j
\end{array}\right].
\label{eqn:CTSystemDynamics}
\end{equation}

Here, $A_b$ is the centroidal momentum matrix which maps generalized velocities to centroidal momentum. Readers can refer to \cite{Orin2013CentroidalDO} for more details.

\subsubsection{Constraints}
The contact constraints (\ref{eqn:ContactConstraint}), (\ref{eqn:NoContactConstraints}) are described in section \ref{subsec:HardContact}. An input limit constraint is also added.

\subsubsection{Cost}
The cost is a quadratic tracking cost to follow a given full state trajectory, including base pose, momentum, and nominal joint positions.

\subsection{Contact-Implicit MPC}
\subsubsection{System Dynamics}
The system states $\mathbf{\tilde{x}} \in \mathbb{R}^{16}$ and inputs $\mathbf{\tilde{u}} \in \mathbb{R}^{13}$ are defined as:

$$
\mathbf{\tilde{x}} = \left[\mathbf{h}_{b}^T, \mathbf{q}_{b}^{T},\mathbf{q}_{j}^{T}\right]^{T}, \mathbf{\tilde{u}} = \left[\mathbf{f}_{c}^T,\mathbf{v}_j^T,
\alpha^T\right]^T.
$$
Here, $\alpha = \left[\alpha_{\text{tangent}}^1,\alpha_{\text{z}}^1,\alpha_{\text{tangent}}^2,\alpha_{\text{z}}^2 \right] \in \mathbb{R}^4$ are the auxiliary input variables that control the tangential parts of the contact force. The other parts of the state and input are consistent with the definitions in the hybrid MPC.

From equation~(\ref{eqn:SoftContactEqn}) we have:
\begin{equation}
\label{normalforce}
    \lambda_{normal}^i = f(D(FK_{i}(\mathbf{q}_{b},\mathbf{q}_{j}))) .
\end{equation}

Here, $FK_{i}(\mathbf{q}_{b},\mathbf{q}_{j})$ is the forward kinematics function that calculates the position of the $i$-th end effector in world frame. $\lambda_{normal}^i$ is the normal part of contact force on the end effector. The tangential parts of the contact force are controlled by the auxiliary variable $\mathbf{\alpha}$. We have:
\begin{gather}
    \lambda_{tan}^i =\alpha_{tan}^i  \cdot \lambda_{normal}^i, \\
    \lambda_{z}^i = \alpha_{z}^i \cdot \lambda_{normal}^i.
\end{gather}

Then, the contact force $\mathbf{f}_c^i$ can be expressed as:
\begin{equation}
    \mathbf{f}_c^i = \lambda_{\text{normal}}^i \mathbf{e}_{\text{normal}}^i + 
\lambda_{\text{tangent}}^i \mathbf{e}_{\text{tangent}}^i + \lambda_{z}^i \mathbf{e}_{\text{z}}^i .
\label{eqn:SoftContactForce}
\end{equation}

Substitute (\ref{eqn:SoftContactForce}) into (\ref{eqn:CTSystemDynamics}), and we can have the system dynamics.

\subsubsection{Constraints} 

The end effector constraints (\ref{eqn:posconstraint}),~(\ref{eqn:velconstraint}) are described in section \ref{subsec:SoftContact}. Additionally, we have a friction cone constraint $-\mu \leq \mathbf{\alpha} \leq \mu$. All these constraints are handled with penalty methods and treated as parts of the cost function.

\subsubsection{Cost}
Like \ref{sec:hybridmpc}, the system cost is a quadratic tracking cost to follow a full state trajectory.

\subsection{Contact Schedule Generation}
The contact schedule used by the Hybrid MPC is generated by thresholding the outputs from contact-implicit MPC. With Eqn. \ref{eqn:normalForce}, we can obtain the normal force trajectories $\lambda_{normal}^i(t)$. The algorithm goes through the force trajectories and add the end effector with $\lambda_{normal}^i(t) > \lambda_{threshold}$ to the active contact set $C(t)$. Here, we use $\lambda_{threshold} = 5$~N. Contacts last shorter than 0.5 second will be neglected.

\subsection{Body control}
The Body Controller consists of two parts: a balancing controller and a yaw controller. The balancing controller is a PID controller cascaded with a PD controller that tracks the desired body leaning angle and velocity. The yaw controller is a PID controller that tracks the yaw orientation in world frame.

\subsection{Arm control}
The arm controller tracks the end effector position and output force at the same time. The reference end effector position is given by:
\begin{equation}
    {}_{B}\mathbf{\tilde{p}^i} = {}_{B}FK_{i}(\mathbf{\tilde{q}}_{j}).
\end{equation}
$\mathbf{{}_{B}\tilde{p}_{ee}^i}$ is the reference end effector position in body frame, and $\mathbf{\tilde{q}}_{j}$
is the reference joint position. We use a task space impedance controller to track the end effector position. The control law is:
\begin{equation}
    {}_{B}\mathbf{f}_{imp} = \mathbf{K}_{p}(\mathbf{{}_{B}\tilde{p}^i} - {}_{B}\mathbf{p}^i) + \mathbf{K}_{d}(0 - {}_{B}\mathbf{\dot{p}}^{i}),
\end{equation}
where $\mathbf{K}_{p}$, $\mathbf{K}_{d}$ are positive definite gain matrices.

Then, the control law used to compute joint torques for the $i$-th  arm is:
\begin{equation}
    \mathbf{\tau^i} = \mathbf{J}^{iT} \left[{}_{B}\mathbf{f}_{imp}^{i} + {}_{B}\mathbf{f}_{c}^{i} \right],
\end{equation}
where $\mathbf{J}^{i}$ is the Jacobian matrix of $i$-th end effector, and $\mathbf{K}_{p}$, $\mathbf{K}_{d}$ are positive definite gain matrices.

\section{Experiments}\label{sec: experiments}

\subsection{Controller Implementation}

The nonlinear optimal control problem is implemented in C++ and solved by the SLQ solver provided by the ETH OCS2~\cite{OCS2} toolbox. The software uses the Eigen3 linear algebra library \cite{eigenweb}. All the dynamics and constraints are implemented with CppAD and can be auto differentiated. 

The controller is deployed on a robot onboard computer with Intel Core i7-1165G7 CPU. The state estimation of the base pose is provided by a T265 tracking camera. Both MPCs have a 1 s planning horizon. The hybrid MPC requires~1.98 ms solve time on average, while the contact-implicit MPC requires 10.99 ms on average. However, to save computation power for other components (navigation, obstacle detection) and ensure software stability, the contact-implicit MPC is restricted to updates at $15$ Hz, and the low-level hybrid MPC updates at $200$ Hz.

%

\subsection{Contact Model}

\begin{figure}[t!h]
    \centering
    \includegraphics[width=\columnwidth]{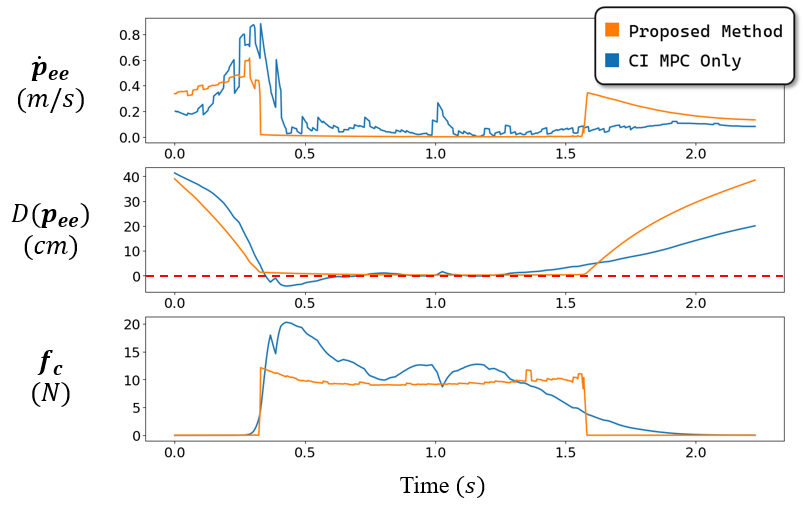}
    \vspace{-5mm}
    \caption{The planned end effector velocity (top), position (middle), and force (bottom) trajectories from the proposed bi-level MPC (orange) and CI MPC (blue).}
    \label{fig:EndEffectorMotion}
    \centering
    \vspace*{-3mm}
\end{figure}
\vspace{-2mm}
In this experiment, we compared the end-effector and force output trajectories generated by a single-level (using only CI-MPC) and bi-level (using Hybrid MPC for trajectory correction) MPC. The time between $0.35$~s and $1.7$~s was identified as a valid contact period. The Hybrid MPC further refined the trajectory based on this constraint. As shown in Fig.~\ref{fig:EndEffectorMotion}, the top plot is the end-effector velocity. During contact, the Hybrid MPC keeps the end-effector velocity close to zero, while the raw output from CI-MPC exhibits noticeable relative sliding. The middle plot displays the distance between the end-effector and the wall. The raw output from CI-MPC has a wall penetration of up to 5 cm, whereas the corrected end-effector trajectory stays precisely on the wall. The bottom plot shows the force output trajectory during this period. Comparing it with the position trajectory reveals that CI-MPC begins outputting force even before the end-effector makes contact with the wall.

It is important to note that the infeasible trajectories produced by the soft CI-MPC can be reduced by tuning contact model parameters. However, to guarantee feasible motion plans online, we find it necessary to apply the hybrid MPC in our framework.
\vspace{-2mm}

\subsection{Disturbance Rejection}

We first test the robot's ability to reject external disturbances by pushing against the wall. The robot was placed 0.5 meters away from the wall and commanded to stay in place. During the test, an operator pushes the robot toward the wall. The time-lapse sequence of the experiment is shown in Fig. \ref{fig:DisturbanceRejectionTimeElapse}(a).

As can be seen, when the robot approached the wall, it stretched its arm and used its end effector to push itself back to its original position. Fig. \ref{fig:DisturbanceRejectionTimeElapse}(b) shows the forces, acceleration, and planned contact sequence from three consecutive trials. The first plot shows the expected force output of the robot's right arm during the experiment. It can be observed that the output quickly reached approximately 20~N when contact occurred. The dominant part of the force came from the output of the hybrid MPC. The task-space impedance controller only has a limited contribution. The second plot shows the magnitude of the robot's acceleration during the process. The peak acceleration marked with red arrows was caused by the disturbance applied by the operator. While the higher peaks resulted from the robot pushing against the wall. After contact ended, the robot's acceleration rapidly decreased.

\subsection{Obstacle Avoidance}

\begin{figure}[t!h]
    \centering
    \includegraphics[width=0.93\columnwidth]{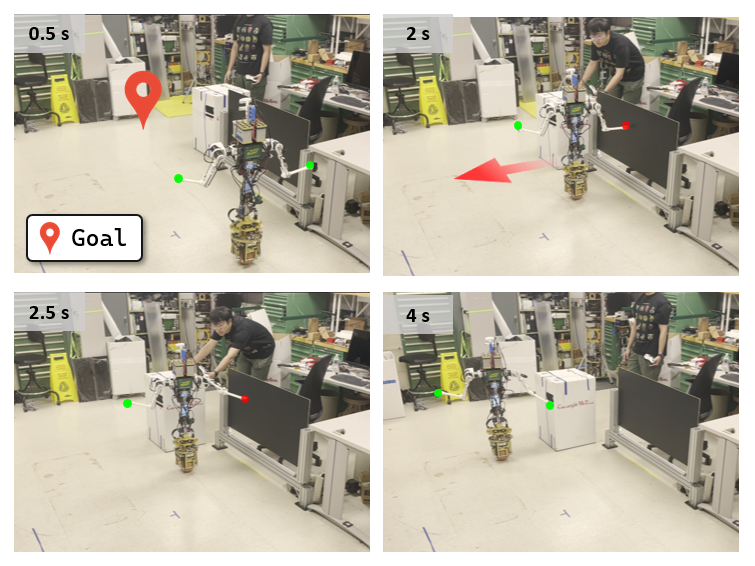}
    \caption{Key frames for the obstacle avoidance experiment. The red highlighted point is the goal location. At T = $2$ s, a obstacle occurred right in front of the robot. The robot pushed the wall to avoid it.}
    \label{fig:Obstacleavoidance}
    \centering
    \vspace*{-3mm}
\end{figure}
\vspace{-2mm}

In this experiment, we demonstrated the robot's capability to quickly avoid obstacles by pushing against a wall. We use an external path planner to provide a reference trajectory to the MPC controller. When an obstacle is detected, the planner will quickly generate a collision-free trajectory. The robot uses a Realsense D435i depth camera to detect obstacles. As shown in Fig.~\ref{fig:Obstacleavoidance}, the robot was commanded to move to a point 3~m ahead. An operator quickly pushed a box to block the robot on its way. Upon detecting the obstacle, our robot successfully pushed against the wall and avoided the obstacle. Note that the acceleration required to avoid the sudden obstacle, in this case, is challenging and dangerous with the IMBD drive wheel alone, making the upper-limb contact force necessary to successfully avoid the obstacle.

\section{Conclusion}\label{sec: conclusion}

In this paper, we proposed a bi-level MPC framework on the CMU shmoobot platform that can utilize non-periodic contacts without predefined contact sequences in locomotion tasks. The proposed framework can provide an optimal trajectory that satisfies hard contact constraints at real-time rates. We validated our approach on a CMU shmoobot, a dual-arm mobile base robot that balances on a ball. The robot shows the ability to utilize contacts to quickly accelerate, decelerate, and avoid dynamic obstacles. The proposed framework can also be deployed on other robotic systems with manipulators. However, in this paper, we only considered contact between two end effectors and vertical walls. Future work could incorporate advanced collision detection libraries to account for whole-body contact. However, we only considered vertical walls. Future work could include surfaces with different orientations, and incorporate legged locomotion into the same framework. Moreover, the algorithm's performance on long-horizon tasks needs to be investigated.





\section*{ACKNOWLEDGMENT}

This work was supported in part by NSF grant CNS-1629757 and Amazon. The authors would like to thank Zhongyu Li, Guanqi He, and Yuntian Zhao for the valuable discussions.


\newpage

\bibliographystyle{IEEEtran}
\bibliography{IEEEabrv,ref.bib}

\end{document}